# Discretized Approximations for POMDP with Average Cost

**Huizhen Yu**
Lab for Information and Decisions
EECS Dept., MIT
Cambridge, MA 02139

**Dimitri P. Bertsekas**
Lab for Information and Decisions
EECS Dept., MIT
Cambridge, MA 02139

## Abstract

In this paper, we propose a new lower approximation scheme for POMDP with discounted and average cost criterion. The approximating functions are determined by their values at a finite number of belief points, and can be computed efficiently using value iteration algorithms for finite-state MDP. While for discounted problems several lower approximation schemes have been proposed earlier, ours seems the first of its kind for average cost problems. We focus primarily on the average cost case, and we show that the corresponding approximation can be computed efficiently using multi-chain algorithms for finite-state MDP. We give a preliminary analysis showing that regardless of the existence of the optimal average cost $J^*$ in the POMDP, the approximation obtained is a lower bound of the liminf optimal average cost function, and can also be used to calculate an upper bound on the limsup optimal average cost function, as well as bounds on the cost of executing the stationary policy associated with the approximation. We show the convergence of the cost approximation, when the optimal average cost is constant and the optimal differential cost is continuous.

## 1 INTRODUCTION

We consider discrete-time infinite horizon partially observable Markov Decision Processes (POMDP) with the state space $\mathcal{S}$, the observation space $\mathcal{Z}$ and the control space $\mathcal{U}$ all being finite. Let $X$ be the set of probability distributions on $\mathcal{S}$, called belief space, and $g_u(s)$ be the per stage cost function. With the average cost criterion, we minimize over the policies $\pi$ the average expected cost $\frac{1}{N} \mathrm{E}^\pi \{ \sum_{t=0}^{N-1} g_{u_t}(s_t) | s_0 \sim x \}$, as $N$ goes to infinity, when the initial state $s_0$ follows the distribution $x$. POMDPs with average cost criterion are substantially more difficult to analyze than with discounted cost. Although there are optimality equations whose solution provides the optimal average cost function and a stationary optimal policy, in general there is no guarantee that a solution exists, and there are no finite computation algorithms to obtain it. Therefore, discretized approximations are computationally appealing as approximate solutions for average cost POMDP, since the problem of finite-state MDPs with average cost is well understood and can be solved with several commonly used algorithms.

We note that a discretization scheme for discounted POMDP that gives a lower approximation was first proposed by (Lovejoy, 1991). It was later improved by (Zhou and Hansen, 2001). There have been no proposals of discretization schemes for average cost POMDP, to our knowledge. A conceptually different alternative to solve approximately average cost POMDP is the finite memory approach (Aberdeen and Baxter, 2002). In this approach, one seeks a policy that is average cost optimal within a class of finite state controllers. The advantage of the finite memory approach is that a suboptimal policy can be learned in a model-free fashion, i.e., with a simulator rather than an explicit transition probability model of the system. By contrast the discretization approaches of Lovejoy, and Zhou and Hansen, as well as ours, require an exact mechanism for generating beliefs/conditional state distributions, as the system is operating.

We have recently become aware of the related work by (Ormoneit and Glynn, 2002) on MDP with continuous state space and average cost. Our POMDP scheme can be viewed as a special case of their general approximation scheme. However, the lower approximation property is special to POMDP, and the corresponding asymptotic convergence results are also different in the two works.

The starting point for our discretization methodology is the discounted problem, for which we introduce a new lower approximation scheme, based on a fictitious "optimistic" controller that receives extra information about the hidden states. The cost of this controller, a lower bound to the optimal cost, can be calculated using finite-state MDP methods, and can be used as an approximate cost-to-go function for a one-step lookahead scheme. We extend our approach to the average cost criterion, where the discretized problem can be solved by multi-chain algorithms for finite-state MDP. We show that the corresponding approximate cost is a lower bound to the optimal liminf average cost function, and can be used to obtain an upper bound to the optimal limsup average cost function, as well as bounds on the cost of the stationary policy associated with the approximation. We show asymptotic convergence of the cost approximation of the discretization scheme, assuming that the optimal average cost is constant and the optimal differential cost is continuous.

The paper is organized as follows. In Section 2, we consider discretized approximations in the discounted case, and introduce a new approximation scheme. We prove asymptotic convergence for two main discretization schemes. In Section 3, we extend discretized approximations to the average cost case, and give an analysis of error bounds and asymptotic convergence. Finally in Section 4, we present experimental results.

Due to space limitations, some of the proofs have been omitted. They can be found in an expanded version of this paper (Yu and Bertsekas, 2004), which also addresses some additional topics, including a general framework for deriving upper and lower approximation schemes for POMDP.

## 2   DISCOUNTED CASE

We introduce a new approximation scheme and summarize known discretized lower approximation schemes for the discounted case. The belief MDPs associated with them will be the basis for the lower approximation schemes in the average cost case. The results obtained here will also be useful there.

In the discounted case, we minimize the discounted cost

$$\mathrm{E}^\pi\{\sum_t \alpha^t g_{u_t}(s_t)|s_0 \sim x\}$$

for a fixed $\alpha \in [0, 1)$. The optimal cost function $J_\alpha^*(x)$ satisfies the Bellman equation

$$J_\alpha^*(x) = (T J_\alpha^*)(x),$$

where

$$(TJ)(x) = \min_{u \in \mathcal{U}} [x'g_u + \alpha \mathrm{E}_z \{J(\phi_u(x, z))\}],$$

and $'$ denotes transpose, $g_u$ denotes the per stage cost vector, and $\phi_u(x, z)$ denotes the conditional distribution of $s_1$ after applying control $u$ and observing $z$.

A few notations for expectations will be used throughout the text. At places where emphasis of the distribution is necessary, we use the symbol $\mathrm{E}_{z|x,u}\{\dots\}$, which should be read as $\sum_z p(z|x, u) \dots$, and is equivalent to the conditional expectation $\mathrm{E}_z\{\dots|x, u\}$.

### 2.1   A NEW INEQUALITY

The optimal cost $J_\alpha^*(\cdot)$ is concave, i.e., for any convex combination $\bar{x} = \sum_i \gamma_i(\bar{x})x_i$, where $\gamma_i(\bar{x}) \geq 0$ and $\sum_i \gamma_i(\bar{x}) = 1$, $J_\alpha^*(\bar{x}) \geq \sum_i \gamma_i(\bar{x})J_\alpha^*(x_i)$. Using this property with $\bar{x} = \phi_u(x, z)$ in the Bellman equation, we have the following inequality that was proposed by (Zhou and Hansen, 2001) for a discretized cost approximation:

$$J_\alpha^*(x) \geq \min_u \left[x'g_u + \alpha \mathrm{E}_{z|x,u}\left\{\sum_i \gamma_i(\phi_u(x, z))J_\alpha^*(x_i)\right\}\right]. \tag{1}$$

We introduce a new inequality, which follows from concavity of $\mathrm{E}_{z|x,u}\{J_\alpha^*(\phi_u(x, z))\}$ in $x$.

**Proposition 1** *For all $x \in X$, $x_i \in X$ and $\gamma_i(x) \geq 0$ such that $x = \sum_i \gamma_i(x)x_i$, $\sum_i \gamma_i(x) = 1$, the optimal cost $J_\alpha^*(x)$ satisfies*

$$J_\alpha^*(x) \geq \min_u \left[x'g_u + \alpha \sum_i \gamma_i(x)\mathrm{E}_{z|x_i,u}\{J_\alpha^*(\phi_u(x_i, z))\}\right]. \tag{2}$$

We present here, however, an alternative proof, that uses the interpretation of a modified process in which there is additional information about the randomness of the initial distribution. This argument has the same spirit as "region-observable-POMDP" (Zhang and Liu, 1997), and can be generalized (Yu and Bertsekas, 2004). Since Prop. 1 implies concavity of $J_\alpha^*(\cdot)$,[1] which is not used in the proof, it can also be used to establish concavity of $J_\alpha^*(\cdot)$ without an induction argument.

**Proof:**   Consider a new process $\mathcal{P}$, otherwise identical to the original POMDP, except that the initial distribution of $s_0$ is generated by a mixture of $m$ distributions $x_i$ marginally identical to $x$. By this we mean that there is a random variable $q$ taking values from 1 to $m$ with

$$p(q = k|x) = \gamma_k(x), \quad p(s_0|q = k) = x_k(s_0).$$

Assume $q$ is not accessible to the controller. The optimal cost for this new process equals $J_\alpha^*(x)$, and is achieved by the policy $\pi^*$ that is optimal in the original POMDP. Denote its action at $x$ by $a$. We have

$$J_\alpha^*(x) = x'g_a + \mathrm{E}\Big\{\mathrm{E}^{\pi^*}\Big\{\sum_{t=1}^\infty \alpha^t g(s_t, u_t)|x, a, z, q\Big\}|x, a\Big\}.$$

---

[1]This is so because $x'g_u = \sum_i \gamma_i(x)x_i'g_u$ and $\min \sum \geq \sum \min$.

Let $\tilde{\phi}_a(x, (z, q))$ be the distribution $p(s_1|x, a, (z, q))$. As $q$ and hence $\tilde{\phi}_a(x, (z, q))$ are inaccessible to $\pi^*$, by the optimality of $J_\alpha^*(\cdot)$, we have that in the last equation

$$\mathrm{E}^{\pi^*}\Big\{\sum_{t=1}^\infty \alpha^t g(s_t, u_t)|x, a, z, q\Big\} \geq \alpha J_\alpha^*(\tilde{\phi}_a(x, (z, q))).$$

Since $\tilde{\phi}_a(x, (z, q)) = \phi_a(x_i, z)$ given $q = i$, it follows that

$$\begin{aligned}
J_\alpha^*(x) &\geq x'g_a + \alpha \mathrm{E}_{(z,q)|x,a}\big\{J_\alpha^*(\tilde{\phi}_a(x, (z, q)))\big\} \\
&= x'g_a + \alpha \sum_i \gamma_i(x)\mathrm{E}_{z|x_i,a}\big\{J_\alpha^*(\phi_a(x_i, z))\big\} \\
&\geq \min_u \big[x'g_u + \alpha \sum_i \gamma_i(x)\mathrm{E}_{z|x_i,u}\big\{J_\alpha^*(\phi_u(x_i, z))\big\}\big].
\end{aligned}$$

$\square$

## 2.2 DISCRETIZED APPROXIMATIONS

We first summarize known lower approximation schemes, and then prove asymptotic convergence for two main schemes corresponding to the inequalities (1) and (2).

### 2.2.1 Approximation Schemes

Let $G = \{x_i\}$ be a finite set of beliefs such that their convex hull is $X$. A simple choice is to discretize $X$ into a regular grid, so we refer to $x_i$ as *grid points*. By choosing different $x_i$ and $\gamma_i(\cdot)$ in the inequalities (1) and (2), we obtain lower cost approximations that are functionally determined by their values at a finite number of beliefs.

**Definition 1 ($\epsilon$-Discretization Scheme)** *Call $(G, \underline{\gamma})$ an $\epsilon$-discretization scheme where $G = \{x_i\}$ is a set of $n$ beliefs, $\underline{\gamma} = (\gamma_1(\cdot), \dots, \gamma_n(\cdot))$ is a convex representation scheme such that $x = \sum_i \gamma_i(x)x_i$ for all $x \in X$, and $\epsilon$ is a scalar characterizing the fineness of the discretization, and defined by*

$$\epsilon = \max_{x \in X} \max_{\substack{x_i \in G \\ \gamma_i(x) > 0}} \|x - x_i\|.$$

Given $(G, \underline{\gamma})$, let $\widetilde{T}_{D_i}, i = 1, 2$, be the associated mappings corresponding to the right-hand sides of inequal-

ities (1) and (2), respectively:

$$\begin{aligned}
(\widetilde{T}_{D_1}J)(x) = \min_u \big[&x'g_u + \\
&\alpha \sum_i \mathrm{E}_{z|x_i,a}\big\{\gamma_i(\phi_u(x, z))\big\}J(x_i)\big],
\end{aligned} \quad (3)$$

$$\begin{aligned}
(\widetilde{T}_{D_2}J)(x) = \min_u \big[&x'g_u + \\
&\alpha \sum_i \gamma_i(x)\mathrm{E}_{z|x_i,u}\big\{J(\phi_u(x_i, z))\big\}\big].
\end{aligned} \quad (4)$$

Associated with these mappings are their unique belief MDPs on the continuous belief space $X$, which we will refer as the *modified belief MDPs*. The optimal cost functions $\tilde{J}_i$ in these modified belief MDPs satisfy, respectively,

$$(\widetilde{T}_{D_i}\tilde{J}_i)(x) = \tilde{J}_i(x) \leq J_\alpha^*(x), \quad \forall x \in X, \ i = 1, 2.$$

Both $\tilde{J}_i$ are functionally determined by their values at a finite number of beliefs, which will be called *supporting points*, and whose set is denoted by $C$. In particular, the function $\tilde{J}_1$ can be computed by solving a corresponding finite-state MDP on $C = G = \{x_i\}$, and the function $\tilde{J}_2$ can be computed by solving a corresponding finite-state MDP on $C = \{\phi_u(x_i, z)|x_i \in G, u \in \mathcal{U}, z \in \mathcal{Z}\}$.[2] The computation can thus be done efficiently by variants of value iteration methods, or linear programming.

Usually $X$ is partitioned into convex regions and beliefs in a region are represented as the convex combinations of its vertices. The function $\tilde{J}_1$ is then piecewise linear on each region, and the function $\tilde{J}_2$ is piecewise linear and concave on each region. To see the latter, let $q(x_i, u) = \mathrm{E}_{z|x_i,u}\{\tilde{J}_2(\phi_u(x_i, z))\}$, and we have $\tilde{J}_2(x) = \min_u[x'g_u + \alpha \sum_s \gamma_i(x)q(x_i, u)]$.

The simplest case for both mappings is when $G$ consists of vertices of the belief simplex, i.e. $G = \{e_s|s \in \mathcal{S}\}$, where $e_s(s) = 1$ and $e_s(s') = 0, s' \neq s, \forall s, s' \in \mathcal{S}$. Denote the corresponding mappings by $\widetilde{T}_{D_i^0}, i = 1, 2$, respectively, i.e.,

$$(\widetilde{T}_{D_1^0}J)(x) = \min_u \big[x'g_u + \alpha \sum_s p(s|x, u)J(e_s)\big], \quad (5)$$

$$\begin{aligned}
(\widetilde{T}_{D_2^0}J)(x) = \min_u \big[&x'g_u + \\
&\alpha \sum_s x(s)\mathrm{E}_{z|s,u}\big\{J(\phi_u(e_s, z))\big\}\big].
\end{aligned} \quad (6)$$

The mapping $\widetilde{T}_{D_1^0}$ is the QMDP approximation, suggested by (Littman, Cassandra, and Kaelbling, 1995), who have shown good results for certain applications. In the belief MDP associated with $\widetilde{T}_{D_1^0}$, the states will be observable after the initial step. In the belief MDP associated with $\widetilde{T}_{D_2^0}$, the *previous* state will be revealed at each stage. One can show that $\widetilde{T}_{D_2^0}$ gives a better approximation than $\widetilde{T}_{D_1^0}$ in both discounted and average cost cases. For the comparison of $\widetilde{T}_{D_i}$ in general, by concavity of $J_\alpha^*$, one can relax the inequality $J_\alpha^* \geq \widetilde{T}_{D_2} J_\alpha^*$ to obtain an inequality of *the same form* as the inequality $J_\alpha^* \geq \widetilde{T}_{D_i} J_\alpha^*$. See (Yu and Bertsekas, 2004) for these details.

---

[2] More precisely, $C = \{\phi_u(x_i, z)|x_i \in G, u \in \mathcal{U}, z \in \mathcal{Z}, \text{ such that } p(z|x_i, u) > 0\}$.

By concatenating mappings we obtain other discretized lower approximations. For example,

$$T \circ \widetilde{T}_{D_i}, \ i = 1, 2; \qquad \widetilde{T}_I \circ \widetilde{T}_{D_2}, \tag{7}$$

where $\widetilde{T}_I$ denotes a region-observable-POMDP type of mapping (Zhang and Liu, 1997). In the concatenated mapping $(\widetilde{T}_I \circ \widetilde{T}_{D_2})$ we only need grid points to be on lower dimensional spaces.

Let $\widetilde{T}$ be any of the above mappings. Its associated modified belief MDP is not necessarily a POMDP model. It is straightforward to show the following,[3] by comparing the $N$-stage optimal cost of the modified MDP to that of the original POMDP. This result also holds for $\alpha = 1$.

**Proposition 2** *Let* $J_0$ *be a concave function on* $X$. *For any* $\alpha \in [0, 1]$,

$$(\widetilde{T}^N J_0)(x) \leq (T^N J_0)(x), \quad \forall x \in X, \ \forall N.$$

### 2.2.2 Asymptotic Convergence

We will now provide a limiting theorem for $\widetilde{T}_{D_1}$ and $\widetilde{T}_{D_2}$ using the uniform continuity property of $J_\alpha^*(\cdot)$. We first give some conventional notations related to policies, to be used throughout the paper. Let $\mu$ be a stationary policy, and $J_\mu$ be its cost. We define the mapping $T_\mu$ by

$$(T_\mu J)(x) = x' g_{\mu(x)} + \alpha E_{z|x,\mu(x)} \{J(\phi_{\mu(x)}(x, z))\},$$

and similarly for any control $u$, we define $T_u$ to be the mapping that has the same single control $u$ in place of $\mu(x)$ in $T_\mu$. Let $\widetilde{T}$ be either $\widetilde{T}_{D_1}$ or $\widetilde{T}_{D_2}$, and similarly let $\widetilde{T}_\mu$ and $\widetilde{T}_u$ correspond to a policy $\mu$ and control $u$, respectively.

The function $J_\alpha^*(x)$ is continuous on $X$. For any continuous function $v(\cdot)$, $E_{z|x,u}\{v(\phi_u(x, z))\}$ is also continuous on $X$. As $X$ is compact, by the uniform continuity of corresponding functions, we have the following lemma.

**Lemma 1** *Let* $v(\cdot)$ *be a continuous function on* $X$. *For any* $\delta > 0$, *there exists* $\bar{\epsilon} > 0$ *such that for any* $\epsilon$-*discretization scheme* $(G, \underline{\gamma})$ *with* $\epsilon \leq \bar{\epsilon}$,

$$|(T_u v)(x) - (\widetilde{T}_u v)(x)| \leq \delta, \ \forall x \in X, \ \forall u \in \mathcal{U},$$

[3]Use induction and concavity, or alternatively an argument similar to the proof of Prop. 1.

where $\widetilde{T}$ *is either* $\widetilde{T}_{D_1}$ *or* $\widetilde{T}_{D_2}$ *associated with* $(G, \underline{\gamma})$.

By Lemma 1, and the standard error bounds $\|\tilde{J} - J^*\|_\infty \leq \frac{\|T\tilde{J} - \tilde{J}\|_\infty}{1-\alpha}$ and $\|J_\mu - \tilde{J}\|_\infty \leq \frac{\|T_\mu \tilde{J} - \tilde{J}\|_\infty}{1-\alpha}$ (see e.g., (Bertsekas, 2001)), we have the following limiting theorem, which states that the lower approximation and the cost of its look-ahead policy, as well as the cost of the "optimal" policy with respect to the modified belief MDP, all converge to the optimal cost of the original POMDP.

**Theorem 1** *Let* $(G_k, \underline{\gamma}_k)$ *be a sequence of* $\epsilon_k$-*discretization schemes with* $\epsilon_k \to 0$ *as* $k \to \infty$. *Let* $\tilde{J}_k$, $\mu_k$ *and* $\tilde{\mu}_k$ *be such that*

$$\tilde{J}_k = \widetilde{T}_k \tilde{J}_k = \widetilde{T}_{k,\tilde{\mu}_k} \tilde{J}_k, \qquad T_{\mu_k} \tilde{J}_k = T \tilde{J}_k,$$

*where* $\widetilde{T}_k$ *is either* $\widetilde{T}_{D_1}$ *or* $\widetilde{T}_{D_2}$ *associated with* $(G_k, \underline{\gamma}_k)$. *Then for any fixed* $\alpha \in [0, 1)$,

$$\tilde{J}_k \to J_\alpha^*, \quad J_{\mu_k} \to J_\alpha^*, \quad J_{\tilde{\mu}_k} \to J_\alpha^*, \quad as \ k \to \infty.$$

## 3 DISCRETIZED APPROXIMATIONS FOR AVERAGE COST CRITERION

In average cost POMDP, the objective is to minimize the average cost $\frac{1}{N} E^\pi \{\sum_{t=0}^{N-1} g(s_t, u_t) | s_0 \sim x_0\}$, as $N$ goes to infinity. For POMDP with average cost, in order that a stationary optimal policy exists, it is sufficient that the following functional equations, in the belief MDP notation,

$$J(x) = \min_u E_{\tilde{x}|x,u} \{J(\tilde{x})\}, \tag{8}$$

$$J(x) + h(x) = \min_{u \in U(x)} [x' g_u + E_{\tilde{x}|x,u} \{h(\tilde{x})\}],$$

$$\text{where} \quad U(x) = \arg\min_u E_{\tilde{x}|x,u} \{J(\tilde{x})\},$$

admit a bounded solution $(J^*(\cdot), h^*(\cdot))$. The stationary policy that obtains the minimum is then optimal with its average cost being $J^*(x)$. However, there are no finite computation algorithms to obtain it. (For a general analysis of POMDP with average cost, see (Fernández-Gaucherand, Arapostathis, and Marcus, 1991) or the survey by (Arapostathis et al., 1993).)

We now extend the application of the discretized approximations to the average cost case. First, note that solving the corresponding average cost problem in the discretized approach is much easier. Let $\widetilde{T}$ be any of the mappings from Eq. (3)-(7) in Section 2.2.1. For its associated modified belief MDP, writing $\bar{g}_u(x)$ for cost per-stage, we have the following average cost optimal-

ity equations:

$$J(x) = \min_u \tilde{\mathrm{E}}_{\tilde{x}|x,u}\{J(\tilde{x})\}, \tag{9}$$

$$J(x) + h(x) = \min_{u \in U(x)} [\bar{g}_u(x) + \tilde{\mathrm{E}}_{\tilde{x}|x,u}\{h(\tilde{x})\}],$$

$$\text{where} \quad U(x) = \arg\min_u \tilde{\mathrm{E}}_{\tilde{x}|x,u}\{J(\tilde{x})\},$$

and we use $\tilde{\mathrm{E}}$ to indicate that the expectation is taken with respect to the distributions $p(\tilde{x}|x, u)$ of the modified MDP, which satisfy

$$p(\tilde{x}|x, u) = 0, \quad \forall (x, u), \forall \tilde{x} \notin C,$$

with $C$ being the finite set of supporting beliefs. There are bounded solutions $(\tilde{J}(\cdot), \tilde{h}(\cdot))$ to the optimality equations (9) for the following reason: Every finite-state MDP problem admits a solution to its average cost optimality equations. Furthermore if $x \notin C$, $x$ is transient and unreachable from $C$, and the next belief $\tilde{x}$ belongs to $C$ under any control $u$ in the modified MDP. It follows that the optimality equations (9) restricted on $\{x\} \cup C$ are the optimality equations for the finite-state MDP with $|C| + 1$ states, so the solution $(\tilde{J}(\bar{x}), \tilde{h}(\bar{x}))$ exists for $\bar{x} \in x \cup C$ with their values on $C$ independent of $x$. This is essentially the algorithm to solve $\tilde{J}(\cdot)$ and $\tilde{h}(\cdot)$ in two stages, and obtain an optimal stationary policy for the modified MDP.

Concerns arise, however, about using any optimal policy for the modified MDP as suboptimal control in the original POMDP. Although all average cost optimal policies behave equally optimally in the asymptotic sense, they do so in the *modified MDP*, in which all the states $x \notin C$ are transient. As an illustration, suppose for the completely observable MDP, the optimal average cost is constant over all states, then at any belief $x \notin C$ any control will have the same asymptotic average cost in the modified MDP corresponding to the QMDP approximation scheme. The situation worsens, if even the completely observable MDP itself has a large number of states that are transient under its optimal policies. We therefore speculate that for the modified MDP, we should aim to compute policies with additional optimality guarantees, relating to their finite-stage behaviors. Fortunately for finite-state MDPs, there are efficient algorithms for computing such policies.

In the following we present the algorithm, after a brief review of the related results for finite-state MDP, and give preliminary analysis of error bounds and asymptotic convergence. We show sufficient conditions for the convergence of cost approximation, assuming that the optimal average cost of the POMDP is constant.

## 3.1 ALGORITHM

We first briefly review related results for *finite-state* MDPs. Since average cost measures the asymptotic behavior of a policy, given two policies having the same average cost, one can incur significantly larger cost in finite steps than the other. The concept of $n$-discount optimality is useful for differentiating between such policies. It is also closely related to Blackwell optimality. A policy $\pi^*$ is $n$-*discount optimal* if its cost in the discounted cases satisfy

$$\limsup_{\alpha \to 1} (1 - \alpha)^{-n}(J_\alpha^{\pi^*}(s) - J_\alpha^\pi(s)) \leq 0, \quad \forall s, \forall \pi.$$

By definition an $(n + 1)$-discount optimal policy is also $k$-discount optimal for $k = -1, 0, \ldots, n$. A policy is called *Blackwell optimal*, if it is optimal for all the discounted problems with discount factor $\alpha \in [\bar{\alpha}, 1)$ for some $\bar{\alpha} < 1$. For finite-state MDPs, a policy is Blackwell optimal if and only if it is $\infty$-discount optimal. By contrast, any $(-1)$-discount optimal policy is average cost optimal.

For any finite-state MDP, there exist stationary average cost optimal policies and furthermore, stationary $n$-discount optimal and Blackwell optimal policies. In particular, there exist functions $J(\cdot), h(\cdot)$ and $w_k(\cdot)$, $k = 0, \ldots, n + 1$, with $w_0 = h$ such that they satisfy the following nested equations:

$$J(s) = \min_{u \in U(s)} \mathrm{E}_{\tilde{s}|s,u}\{J(\tilde{s})\}, \tag{10}$$

$$J(s) + h(s) = \min_{u \in U_{-1}(s)} [g_u(s) + \mathrm{E}_{\tilde{s}|s,u}\{h(\tilde{s})\}],$$

$$w_{k-1}(s) + w_k(s) = \min_{u \in U_{k-1}(s)} \mathrm{E}_{\tilde{s}|s,u}\{w_k(\tilde{s})\},$$

where

$$U_{-1}(s) = \arg\min_{u \in U(s)} \mathrm{E}_{\tilde{s}|s,u}\{J(\tilde{s})\},$$

$$U_0(s) = \arg\min_{u \in U_{-1}(s)} [g_u(s) + \mathrm{E}_{\tilde{s}|s,u}\{h(\tilde{s})\}],$$

$$U_k(s) = \arg\min_{u \in U_{k-1}(s)} \mathrm{E}_{\tilde{s}|s,u}\{w_k(\tilde{s})\}.$$

Any stationary policy that attains the minimum in the right-hand sides of the equations in (10) is an $n$-discount optimal policy.

For finite-state MDPs, a stationary $n$-discount optimal policy not only exists, but can also be efficiently computed by multi-chain algorithms. Furthermore, in order to obtain a Blackwell optimal policy, which is $\infty$-discount optimal, it is sufficient to compute a $(N - 2)$-discount optimal policy, where $N$ is the number of states in the finite-state MDP. We refer readers to (Puterman, 1994) Chapter 10, especially Section 10.3 for details of the algorithm as well as theoretical analysis.

This leads to the following algorithm for computing an $n$-discount optimal policy for the *modified* MDP defined on the continuous belief space. We first solve the average cost problem on $C$, then determine optimal controls on transient states $x \notin C$. Note there are no conditions (such as unichain) at all on this modified belief MDP.

**The algorithm solving the modified MDP**

1. Compute an $n$-discount optimal solution for the finite-state MDP problem associated with $C$. Let $\tilde{J}^*(x_i)$, $\tilde{h}(x_i)$, and $\tilde{w}_k(x_i)$, $k = 1, \ldots, n+1$, with $x_i \in C$, be the corresponding functions obtained that satisfy Eq. (10) on $C$.

2. For any belief $x$, let the control set $U_{n+1}$ be computed at the last step of the sequence of optimizations:

$$U_{-1} = \arg\min_u \tilde{\mathrm{E}}_{x_i|x,u}\{\tilde{J}^*(x_i)\},$$

$$U_0 = \arg\min_{u \in U_{-1}} [\bar{g}_u(x) + \tilde{\mathrm{E}}_{x_i|x,u}\{\tilde{h}(x_i)\}],$$

$$U_k = \arg\min_{u \in U_{k-1}} \tilde{\mathrm{E}}_{x_i|x,u}\{\tilde{w}_k(x_i)\}, \ 1 \le k \le n+1.$$

Let $u$ be any control in $U_{n+1}$, and let $\tilde{\mu}^*(x) = u$. Also if $x \notin C$, define

$$\tilde{J}^*(x) = \tilde{\mathrm{E}}_{x_i|x,u}\{\tilde{J}^*(x_i)\},$$

$$\tilde{h}(x) = \bar{g}_u(x) + \tilde{\mathrm{E}}_{x_i|x,u}\{\tilde{h}(x_i)\} - \tilde{J}^*(x).$$

With the above algorithm we obtain an $(n-1)$-discount optimal policy for the modified MDP. When $n = |C| - 1$, we obtain an $\infty$-discount optimal policy for the modified MDP,[4] since the algorithm essentially computes a Blackwell optimal policy for every finite-state MDP restricted on $\{x\} \cup C$. Thus, for the *modified MDP*, for any other policy $\pi$, and any $x \in X$,

$$\limsup_{\alpha \to 1} (1-\alpha)^{-n}(\tilde{J}_\alpha^{\tilde{\mu}^*}(x) - \tilde{J}_\alpha^\pi(x)) \le 0, \quad \forall n \ge -1.$$

It is also straightforward to see that

$$\tilde{J}^*(x) = \lim_{\alpha \to 1}(1-\alpha)\tilde{J}_\alpha^*(x), \quad \forall x \in X, \qquad (11)$$

where $\tilde{J}_\alpha^*(x)$ are the optimal discounted costs for the modified MDP, and the convergence is uniform over $X$, since $\tilde{J}_\alpha^*(x)$ and $\tilde{J}^*(x)$ are piecewise linear interpolations of the function values on a finite set of beliefs.

---

[4]Note that $\infty$-discount optimality and Blackwell optimality are equivalent for finite-state MDPs, however, they are not equivalent in the case of a continuous state space. In the modified MDP, although for each $x$ there exists an $\alpha(x) \in (0,1)$ such that $\tilde{\mu}^*(x)$ is optimal for all $\alpha$-discounted problems with $\alpha(x) \le \alpha < 1$, we may have $\sup_x \alpha(x) = 1$ due to the continuity of the belief space.

## 3.2   ANALYSIS OF ERROR BOUNDS

We now show how to bound the optimal average cost of the original POMDP, and how to bound the cost of executing the suboptimal policy, that is optimal to the modified MDP, in the original POMDP.

Let $V_N^\pi(x) = \mathrm{E}^\pi\{\sum_{t=0}^{N-1} \bar{g}_{u_t}(x_t)|x_0 = x\}$ be the $N$-stage cost of a non-randomized policy $\pi$, which can be non-stationary, in the original POMDP. Let

$$J_-^*(x) = \inf_\pi \liminf_{N \to \infty} \frac{1}{N}V_N^\pi(x), \ J_+^*(x) = \inf_\pi \limsup_{N \to \infty} \frac{1}{N}V_N^\pi(x).$$

It is straightforward to show[5] that $\tilde{J}^*(x) \le J_+^*(x)$, $\forall x \in X$. We now show that $\tilde{J}^*(x) \le J_-^*(x)$, $\forall x \in X$.

**Proposition 3** *The optimal average cost function $\tilde{J}^*(x)$ of the modified MDP satisfies*

$$\tilde{J}^*(x) \le J_-^*(x), \quad \forall x \in X.$$

**Proof:**   Let $V_N^*(x)$ and $\tilde{V}_N^*(x)$ be the optimal $N$-stage cost function of the original POMDP, and of the modified belief MDP, respectively. By Prop. 2 in Section 2.2.1, we have $\tilde{V}_N^*(x) \le V_N^*(x)$, $\forall N$. Thus $\tilde{J}^*(x) = \liminf_{N \to \infty} \frac{1}{N}\tilde{V}_N^*(x) \le \liminf_{N \to \infty} \frac{1}{N}V_N^*(x) \le J_-^*(x)$. $\square$

Next we give a simple upper bound on $J_+^*(\cdot)$.

**Theorem 2** *The optimal liminf and limsup average cost functions satisfy*

$$\tilde{J}^*(x) \le J_-^*(x) \le J_+^*(x) \le \max_{\bar{x} \in C} \tilde{J}^*(\bar{x}) + \delta,$$

*where*   $\delta = \max_{x \in X} \left[ (T\tilde{h})(x) - \tilde{J}^*(x) - \tilde{h}(x) \right],$

*and $\tilde{J}^*(x)$, $\tilde{h}(x)$ and $C$ are defined as in the modified MDP.*

This statement is a consequence of the following lemma, whose proof, omitted here, follows by bounding the expected cost per stage in the summation of the $N$-stage cost.

**Lemma 2** *Let $J(x)$ and $h(x)$ be any bounded functions on $X$, and $\mu$ be any stationary policy. Define*

---

[5]Since in the discounted case the corresponding lower approximation satisfies $\tilde{J}_\alpha^*(x) \le J_\alpha^*(x)$, by Eq. (11) and a Tauberian theorem, we have for the approximate average cost

$$\tilde{J}^*(x) = \lim_{\alpha \to 1}(1-\alpha)\tilde{J}_\alpha^*(x) \le \liminf_{\alpha \to 1}(1-\alpha)J_\alpha^*(x)$$
$$\le \inf_\pi \limsup_{N \to \infty} \frac{1}{N}V_N^\pi(x) = J_+^*(x).$$

constants $\delta^+$ and $\delta^-$ by

$$\delta^+ = \max_{x \in X} \left[ \bar{g}_{\mu(x)}(x) + \mathrm{E}_{\tilde{x}|x,\mu(x)}\{h(\tilde{x})\} - J(x) - h(x) \right],$$

$$\delta^- = \min_{x \in X} \left[ \bar{g}_{\mu(x)}(x) + \mathrm{E}_{\tilde{x}|x,\mu(x)}\{h(\tilde{x})\} - J(x) - h(x) \right].$$

Then $V_N^\mu(x)$, the $N$-stage cost of executing policy $\mu$, satisfies

$$\beta^-(x) + \delta^- \leq \liminf_{N \to \infty} \frac{1}{N} V_N^\mu(x)$$
$$\leq \limsup_{N \to \infty} \frac{1}{N} V_N^\mu(x) \leq \beta^+(x) + \delta^+, \ \forall x \in X,$$

where $\beta^+(x)$, $\beta^-(x)$ are defined by

$$\beta^+(x) = \max_{\bar{x} \in \mathcal{D}_x^\mu} J(\bar{x}), \quad \beta^-(x) = \min_{\bar{x} \in \mathcal{D}_x^\mu} J(x),$$

and $\mathcal{D}_x^\mu$ denotes the set of beliefs reachable under policy $\mu$ from $x$.

Let $\tilde{\mu}^*$ be the stationary policy that is optimal for the modified MDP. We can use Lemma 2 to bound the liminf and limsup average cost of $\tilde{\mu}^*$ in the original POMDP. For example, if the optimal average cost $J_{MDP}^*$ of the completely observable MDP problem equals the constant $\lambda^*$ over all states, then we also have $\tilde{J}^*(x) = \lambda^*$, $\forall x \in X$, for this modified MDP. The cost of executing the policy $\tilde{\mu}^*$ in the original POMDP can therefore be bounded by

$$\lambda^* + \delta^- \leq \liminf_{N \to \infty} \frac{1}{N} V_N^{\tilde{\mu}^*}(x)$$
$$\leq \limsup_{N \to \infty} \frac{1}{N} V_N^{\tilde{\mu}^*}(x) \leq \lambda^* + \delta^+.$$

The quantities $\delta^+$ and $\delta^-$ can be hard to calculate exactly in general, since $\tilde{J}^*(\cdot)$ and $\tilde{h}(\cdot)$ obtained from the modified MDP are piecewise linear functions. The bounds may also be loose. On the other hand, these functions may indicate the structure of the original problem, and help us to refine the discretization scheme in the approximation.

## 3.3   ANALYSIS OF ASYMPTOTIC CONVERGENCE

Let $(G, \underline{\gamma})$ be an $\epsilon$-discretization scheme, and $\tilde{J}_\epsilon$ and $\tilde{J}_{\alpha,\epsilon}$ be the optimal average cost and discounted cost, respectively, in the modified MDP associated with $(G, \underline{\gamma})$ and either $\widetilde{T}_{D_1}$ or $\widetilde{T}_{D_2}$. Recall that in the discounted case (Theorem 1) for a fixed discount factor $\alpha$, we have asymptotic convergence to optimality:

$$\lim_{\epsilon \to 0} \tilde{J}_{\alpha,\epsilon}(x) = J_\alpha^*(x).$$

We now address the question whether $\tilde{J}_\epsilon \to J^*(x)$, as $\epsilon \to 0$, when $J^*(x) = J_-^*(x) = J_+^*(x)$ exists.

This question of asymptotic convergence under the average cost criterion is hard to tackle for a couple of reasons. First of all, it is not clear when $J^*(x)$ exists. (Fernández-Gaucherand, Arapostathis, and Marcus, 1991) have shown that under certain conditions, (such as the condition that $|J_\alpha^*(x) - J_\alpha^*(\bar{x})|$ is bounded for all $\alpha \in [0, 1)$, and its relaxed variants,) the optimal average cost $J^*(x)$ exists and equals a constant $\lambda^*$ over $X$, and furthermore

$$\lambda^* = \lim_{\alpha \to 1} (1 - \alpha) J_\alpha^*(x), \quad \forall x \in X. \tag{12}$$

However, even when Eq. (12) holds, in general we have

$$\lim_{\epsilon \to 0} \tilde{J}_\epsilon(x) = \lim_{\epsilon \to 0} \lim_{\alpha \to 1} (1 - \alpha) \tilde{J}_{\alpha,\epsilon}(x)$$
$$\neq \lim_{\alpha \to 1} \lim_{\epsilon \to 0} (1 - \alpha) \tilde{J}_{\alpha,\epsilon}(x) = \lambda^*.$$

To ensure that $\tilde{J}_\epsilon \to \lambda^*$, we therefore need stronger conditions than those that guarantee the existence of $\lambda^*$. We now show that a sufficient condition is the continuity of the optimal differential cost $h^*(\cdot)$.

**Theorem 3** *Suppose the average cost optimality equations (8) admit a bounded solution $(J^*(x), h^*(x))$ with $J^*(x)$ equal to a constant $\lambda^*$. Then, if the differential cost $h^*(x)$ is continuous on $X$, we have*

$$\lim_{\epsilon \to 0} \tilde{J}_\epsilon(x) = \lambda^*, \quad \forall x \in X,$$

*and the convergence is uniform, where $\tilde{J}_\epsilon$ is the optimal average cost function for the modified MDP corresponding to either $\widetilde{T}_{D_1}$ or $\widetilde{T}_{D_2}$ with an associated $\epsilon$-discretization scheme $(G, \underline{\gamma})$.*

**Proof:** Let $\tilde{\mu}_\epsilon^*$ be the optimal policy for the modified MDP associated with an $\epsilon$-discretization scheme. Let $\widetilde{T}$ be the mapping corresponding to the modified MDP, defined by $(\widetilde{T}v)(x) = \min_u[\bar{g}_u(x) + \widetilde{\mathrm{E}}_{\tilde{x}|x,u}\{v(\tilde{x})\}]$. Since $h^*(x)$ is continuous on $X$, by Lemma 1 in Section 2.2.2, we have that for any $\delta > 0$, there exists $\bar{\epsilon} > 0$ such that for all $\epsilon$-discretization schemes with $\epsilon < \bar{\epsilon}$,

$$|(T_{\tilde{\mu}_\epsilon^*} h^*)(x) - (\widetilde{T}_{\tilde{\mu}_\epsilon^*} h^*)(x)| \leq \delta. \tag{13}$$

We now apply the result of Lemma 2 in the modified MDP with $J = \lambda^*$, $h = h^*$, and $\mu = \tilde{\mu}_\epsilon^*$. That is, by the same argument as in Lemma 2, we have

$$\tilde{J}_\epsilon(x) = \liminf_{N \to \infty} \frac{1}{N} \tilde{V}_N^{\tilde{\mu}_\epsilon^*}(x) \geq \lambda^* + \eta, \ \forall x \in X,$$

where $\eta = \min_{x \in X} \left[ (\widetilde{T}_{\tilde{\mu}_\epsilon^*} h^*)(x) - \lambda^* - h^*(x) \right]$. Since

$$\lambda^* + h^*(x) = (Th^*)(x) \leq (T_{\tilde{\mu}_\epsilon^*} h^*)(x),$$

and $|(T_{\tilde{\mu}_\epsilon^*} h^*)(x) - (\widetilde{T}_{\tilde{\mu}_\epsilon^*} h^*)(x)| \leq \delta$ by Eq. (13), we have

$$(\widetilde{T}_{\tilde{\mu}_\epsilon^*} h^*)(x) - \lambda^* - h^*(x) \geq -\delta.$$

Hence $\eta \geq -\delta$, and $\tilde{J}_\epsilon(x) \geq \lambda^* - \delta$ for all $\epsilon \leq \bar{\epsilon}$, and $x \in X$, which proves the uniform convergence of $\tilde{J}_\epsilon$ to $\lambda^*$. $\qquad \square$

Note that the inequality $\tilde{J}_\epsilon \leq J^*$ is crucial in the preceding proof. Note also that the proof does not generalize to the case when $J^*(x)$ is not constant. A fairly strong sufficient condition that guarantees the existence of a constant $J^*$ and a continuous $h^*$ is that $J^*_\alpha(x)$ is equicontinuous on $X$ for all $\alpha \in [0, 1)$. (For a proof see (Ross, 1968) or Theorem 6.3 (iv) in (Arapostathis et al., 1993)).

## 4    PRELIMINARY EXPERIMENTS

We demonstrate our approach on a set of toy problems: **Paint**, **Bridge-repair**, and **Shuttle**. The sizes of the problems are summarized in Table 1. Their descriptions and parameters are as specified in A. Cassandra's POMDP File Repository (http://cs.brown.edu/research/ai/pomdp/examples), and we define costs to be negative rewards when a problem has a reward model.

Table 1: Sizes of Problems $|\mathcal{S}| \times |\mathcal{U}| \times |\mathcal{Z}|$

| Paint | Bridge | Shuttle |
|---|---|---|
| $4 \times 4 \times 2$ | $5 \times 12 \times 5$ | $8 \times 3 \times 5$ |

We used some simple grid patterns. One pattern, referred to as **k-E**, consists of $k$ grid points on each edge, in addition to the vertices of the belief simplex. Another pattern, referred to as **n-R**, consists of $n$ randomly chosen grid points, in addition to the vertices of the simplex. The combined pattern is referred to as k-E+n-R. Thus the grid pattern for QMDP approximation is 0-E, for instance, and 2-E+10-R is a combined pattern. The grid pattern then induces a partition of the belief space and a convex representation (interpolation) scheme, which we kept implicitly and computed by linear programming on-line.

The algorithm for solving the modified finite-state MDP was implemented by solving a system of linear equations for each policy iteration. This may not be the most efficient way. No higher than 5-discount optimal policies were computed, when the number of supporting points became large.

Figure 1 shows the average cost approximation of $\tilde{T}_{D_1}$ and $\tilde{T}_{D_2}$ with a few grid patterns for the problem Paint. In all cases we obtained a constant average cost for the modified MDP. The horizontal axis is labeled by the grid pattern, and the vertical axis is the approximate cost. The red curve is obtained by $\tilde{T}_{D_1}$, and the blue curve $\tilde{T}_{D_2}$. As will be shown below, the approximation obtained by $\tilde{T}_{D_2}$ with 3-E is already near optimal. The policies generated by $\tilde{T}_{D_2}$ are not always better, however. We also notice, as indicated by the drop in the curves when using grid pattern 4-E, that the improvement of cost approximation does not solely depend on the number of grid points, but also on where they are positioned.

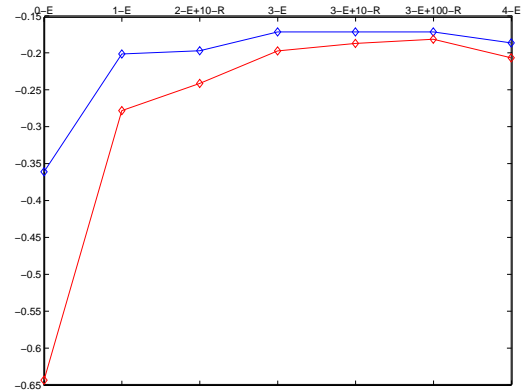

Figure 1: Average Cost Approximation for Problem Paint Using Various Grid Patterns. Blue: $\tilde{T}_{D_2}$, Red: $\tilde{T}_{D_1}$.

In Table 2 we summarize the cost approximations obtained (column **LB**) and the simulated cost of the policies (column **S. Policy**) for the three problems. The approximation schemes obtaining **LB** values in Table 2, as well as the policies simulated, are listed in Table 3. The column **N. UB** shows the numerically computed upper bound of the optimal – we calculate $\delta$ in Theorem 2 by sampling the values of $(T\tilde{h})(x) - \tilde{h}(x) - \tilde{J}(x)$ at hundreds of beliefs generated randomly and taking the maximum over them. Thus the **N. UB** values are under-estimates of the exact upper bound. For both Paint and Shuttle the number of trajectories simulated is 160, and for Bridge 1000. Each trajectory has 500 steps starting from the same belief. The first number in **S. Policy** in Table 2 is the mean over the average cost of simulated trajectories, and the standard error listed as the second number is estimated from bootstrap samples – we created 100 pseudo-random samples by sampling from the empirical distribution of the original sample and computed the standard deviation of the mean estimator over these 100 pseudo-random samples.

As shown in Table 2, we find that some policy from the discretized approximation with very coarse grids can already be comparable to the optimal. This is verified by simulating the policy (**S. Policy**) and comparing its average cost against the lower bound of the optimal

Table 2: Average Cost Approximations and Simulated
Average Cost of Policies

| Problem | LB | N. UB | S. Policy |
|---------|-----|-------|-----------|
| Paint | $-0.170$ | -0.052 | $-0.172$ $\pm0.002$ |
| Bridge | 241.798 | 241.880 | 241.700 $\pm1.258$ |
| Shuttle | $-1.842$ | $-1.220$ | $-1.835$ $\pm0.007$ |

Table 3: Approximation Schemes in LB and Simulated
Policies in Table 2

| Problem | LB | S. Policy |
|---------|-----|-----------|
| Paint | $T_{D_2}$ w/ 3-E | $T_{D_1}$ w/ 1-E |
| Bridge | $T_{D_2}$ w/ 0-E | $T_{D_2}$ w/ 0-E |
| Shuttle | $T_{D_{1,2}}$ w/ 2-E | $T_{D_1}$ w/ 2-E |

(**LB**), which in turn shows that the lower approxima-
tion is near optimal.

We find that in some cases the upper bounds may
be too loose to be informative. For example, in the
problem Paint we know that there is a simple policy
achieving zero average cost, therefore a near-zero up-
per bound does not tell much about the optimal. In
the experiments we also observe that an approxima-
tion scheme with more grid points does not necessarily
provide a better upper bound of the optimal.

## 5    CONCLUSION

In this paper we have proposed a discretized lower
approximation approach for POMDP with average
cost. We have shown that the approximations can be
computed efficiently using multi-chain algorithms for
finite-state MDP, and they can be used for bounding
the optimal liminf and limsup average cost functions,
as well as generating suboptimal policies. Thus, like
the finite state controller approach, our approach also
bypasses the difficult analytic questions such as the
existence of bounded solutions to the average cost op-
timality equations. We have also introduced a new
lower approximation scheme for both discounted and
average cost cases, and shown asymptotic convergence
of two main approximation schemes in the average cost
case under certain conditions.

**Acknowledgements**

This work is supported by NSF Grant ECS-0218328.
We thank Leslie Kaelbling for helpful discussions.